\documentclass[sigconf]{acmart}

\AtBeginDocument{%
  }

\setcopyright{acmlicensed}
\copyrightyear{2024}
\acmYear{2024}
\acmDOI{XXXXXXX.XXXXXXX}

\acmConference[KDD24]{The First Workshop on AI Behavioral Science}{August 26, 2024}{Barcelona, Spain}

\begin{document}

\title{A Comparison of Large Language Model and Human Performance on Random Number Generation Tasks}

\author{Rachel M Harrison}
\email{rae@ophiuchus.ai}
\orcid{0009-0007-2821-2303}
\affiliation{%
  \institution{Ophiuchus LLC}
  \city{Dover}
  \state{DE}
  \country{USA}
}


\begin{abstract}
Random Number Generation Tasks (RNGTs) are used in psychology for examining how humans generate sequences devoid of predictable patterns.
By adapting an existing human RNGT for an LLM-compatible environment, this preliminary study tests whether ChatGPT-3.5, a large language model (LLM) trained on human-generated text, exhibits human-like cognitive biases when generating random number sequences.
Initial findings indicate that ChatGPT-3.5 more effectively avoids repetitive and sequential patterns compared to humans, with notably lower repeat frequencies and adjacent number frequencies.
Continued research into different models, parameters, and prompting methodologies will deepen our understanding of how LLMs can more closely mimic human random generation behaviors, while also broadening their applications in cognitive and behavioral science research.
\end{abstract}




\maketitle

\section{Introduction}



Randomness is commonly defined as the absence of predictability or order \cite{wikipedia2024randomness}, and random number generation tasks (RNGTs) have been used in both psychological research \cite{jahanshahi2006random, towse1998analyzing, jahanshahi2000role} and technological applications \cite{barker2007recommendation, l1999good} to assess and utilize randomness.
These tasks involve generating a sequence of numbers that cannot be described by a sequence shorter than itself, 
which requires the avoidance of recognizable patterns and defined algorithms in order to 
be deemed as random as possible 
\cite{l1992testing}. True randomness is incredibly hard to generate artificially \cite{yuan2015intrinsic}, and most computer-generated random number generations (RNGs) employed in these tasks are actually pseudorandom rather than truly random \cite{l1992testing, dicarlo2012random}. Pseudorandom numbers are generated using algorithms that can produce long sequences of apparently random results, which are entirely determined by an initial value known as a seed. While these pseudorandom numbers appear unpredictable and successfully pass many statistical tests for randomness, they are not genuinely random because their generation is algorithmically determined and can theoretically be reproduced if the seed value is known \cite{l1992testing, dicarlo2012random}.




In psychological contexts, RNGTs are often used to study executive functions, particularly those related to the frontal lobes of the brain, such as inhibition, cognitive flexibility, and the updating and monitoring of information \cite{miyake2000unity, baddeley1996exploring, baddeley1998random, jahanshahi2006random, jahanshahi1998effects, jahanshahi1998left}.
These tasks, which require individuals to generate numbers in a sequence perceived as random, are intrinsically challenging due to the cognitive biases and predictable patterns that typically govern human thought processes.
Historically, studies have shown that humans are suboptimal random number generators, tending to avoid repetition and favor sequential ordering, thereby deviating significantly from true randomness \cite{tune1964response, ginsburg1994random, miyake2000unity}.
The complexity of RNGTs makes them valuable for investigating various aspects of cognitive function and psychopathology, as they require multiple executive processes to work in coordination \cite{peters2007random}.

Faced with the contrasting capabilities of computers and humans in generating random number sequences, large language models (LLMs) like OpenAI's ChatGPT \cite{brown2020language} provide new avenues for the implementation of artificial intelligence (AI) into psychological and behavioral research. Based in the transformer architecture, a type of deep neural network that employs an attention mechanism to process vast amounts of sequential data efficiently and in parallel \cite{vaswani2017attention}, these models excel at natural language processing tasks like text summarization, translation, and question answering \cite{radford2019language}.
Having been trained on vast datasets composed of diverse human-generated texts, these models are inherently imbued with an extensive but implicit understanding of human language, cognition, and social norms \cite{devlin2018bert, goertzel2023generative, grossmann2023ai}. As such, LLMs have the ability to mimic complex human cognitive processes and produce outputs that reflect various human biases and behaviors, despite being fundamentally algorithmic in their operation \cite{brown2020language}.

Recent research underscores the capability of LLMs to emulate complex human cognitive and social behaviors \cite{ke2024exploring}, suggesting a potential overlap with the cognitive processes involved in RNGTs. For instance, studies have highlighted LLMs’ proficiency in perceptual processing, complex problem-solving, decision-making, and creativity \cite{marjieh2023large, orru2023human, hagendorff2023human, stevenson2022putting}. These capabilities indicate that LLMs might not only mimic the surface structure of human language but also the deeper cognitive processes that guide human behavior, potentially including those involved in generating random sequences.

Unlike traditional computational methods that generate numbers via deterministic algorithms, LLMs process and generate sequences in a way that could potentially mirror human randomness, marked by inherent biases and imperfections.
This human-generated training data provides a unique opportunity to study how generative AI (GenAI) might both replicate and diverge from human cognitive biases in tasks that test randomness and decision-making \cite{vaswani2017attention, binz2023using}. Exploring these interactions not only advances our understanding of LLMs themselves, but also provides a novel platform for further analyzing and understanding certain aspects of human linguistic and cognitive skills \cite{goertzel2023generative, sartori2023language}.

This study aims to investigate these possibilities by comparing the performance of LLMs against human performance in RNGTs. We hypothesize that the training data of LLMs, inherently reflective of human behavior, may influence their ability to generate random sequences in ways that are distinct yet subtly human-like. To assess this, we test ChatGPT-3.5 on a simulated RNGT modeled after an existing study conducted with human participants \cite{figurska2008humans}, and utilize common metrics such as the mean frequency of digits and the relative frequencies of consecutive digit pairs to determine whether the model exhibits randomness characteristics that are comparable to or divergent from those most often seen in humans.

\section{Methods}


This study aims to replicate the RNGTs outlined in \cite{persaud2005humans} and \cite{figurska2008humans} using contemporary LLMs. For this project, we focus exclusively on the use of OpenAI's ChatGPT model due to its extensive body of research and status as one of the most advanced and widely recognized models in the field of AI-driven language generation \cite{vogels2023majority, wu2023brief, liu2023summary}.

Specifically, we employ the \texttt{gpt-3.5-turbo-0125} model via OpenAI API calls using the \texttt{chat completion} API to communicate with the model. To ensure that the model's outputs reflect its standard operational parameters, all API calls were made using the default settings as outlined in the official documentation\footnote{\url{https://platform.openai.com/docs/api-reference/chat}}.



In order to accurately simulate the conditions of the preexisting human studies and adapt them for LLMs, we crafted a specific user prompt based on the verbal instructions used in the referenced studies. The original verbal prompt directed participants to "Continue generating and dictating a sequence of random numbers, using the digits 0–9, until you would like to stop" \cite{persaud2005humans}.
Participants were not provided with any definition of "random" \cite{persaud2005humans} or were instructed to choose numbers "in the way they perceived as random" \cite{figurska2008humans}, without the opportunity to ask clarifying questions.

Adjustments to this prompt were necessary to accommodate the operational characteristics of LLMs. During preliminary tests, the direct application of the original instructions resulted in sequences too short for robust analysis. To address this, we modified the prompt to specify a target length for the sequence:

\texttt{Continue generating and dictating a sequence of random numbers, using the digits 0-9, until you reach \{sample\} digits.}

Here, \texttt{\{sample\}} is a random variable drawn from a normal distribution with a mean of $269$ and a standard deviation of $325$ for each sequence generation request, i.e.
\[
    \texttt{sample} \sim \mathcal{N}(269, 325^2).
\]

We also imposed restrictions on generated sequence length to further ensure compliance, with sequences being no fewer than $2$ digits and no more than $922$ digits.
This method of sequence generation ensures that sequence lengths align as much as possible with those observed in human participants~\cite{figurska2008humans}.

To ensure that the results of our study are statistically significant, we collected a total of $10,000$ responses from the \texttt{gpt-3.5-turbo-0125} model.
This dataset far exceeds the typical sample sizes obtainable in human studies (e.g. 37 participants in the comparison study \cite{figurska2008humans}) where logistical complexities and participant availability impose significant constraints. In contrast, the scalability of data collection with LLMs is predominantly restricted only by API rate limits and the computational budget, thus enabling the generation of substantially more data that can be used for comprehensive and statistically significant analysis.


Model responses were collected and processed to remove all non-numeric characters. All operations were executed in a Jupyter notebook environment using Python 3.8 on a consumer-grade laptop. Full source code and generated datasets can be found at:
\url{https://github.com/paxnea/genAI-rngt}.

\section{Results}

This section presents the analysis of $10,000$ sequences generated by ChatGPT in comparison to human-generated data and theoretical expectations of truly random sequences. While ChatGPT demonstrates a higher level of apparent randomness than human outputs, it still falls short of achieving perfect randomness. Notably, the ChatGPT-generated sequences exhibit a marked avoidance of consecutive duplicate digits and adjacent number sequences, both increasing and decreasing, at rates lower than those observed in both human and pseudorandom generation.



The average length of the generated sequences was $308 (\pm 711)$, closely resembling the length observed in human-generated sequences from the reference study \cite{figurska2008humans}, which reported averages of $269 (\pm 325)$ for voluntary generations. While this alignment is likely largely attributable to the structured prompt design that explicitly dictates the length constraints rather than an inherent characteristic of the LLM's behavior, it is nevertheless listed here for full analytical context.



The mean value of all digits across the generated samples was $4.492 (\pm 0.070)$, as compared to the $4.537$ reported for voluntarily generated sequences in the human study \cite{figurska2008humans}, with both datasets thus indicating a close approximation to expected values under ideally random conditions ($4.5$).


\subsubsection*{Pattern Frequencies}

\begin{figure}
    \includegraphics[width=\linewidth]{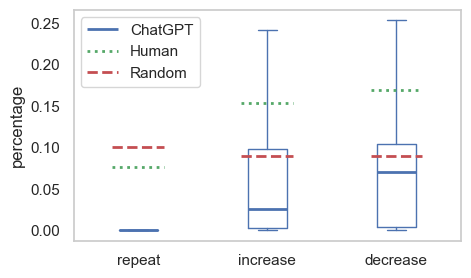}
    \caption{Comparison of pattern frequencies between ChatGPT, humans, and uniformly random distribution.}
    \Description{Pattern frequencies of ChatGPT, human participants, and the uniformly random distribution.}
    \label{fig:patterns_frequency}
\end{figure}

Informed by framework utilized for RNGT analysis in the comparison study \cite{figurska2008humans}, we assessed the 'randomness' of the sequences through pattern-specific metrics like \textit{repeat frequency}, \textit{increase frequency}, and \textit{decrease frequency}. The comparison of these pattern metrics between GPT generations, human participants~\cite{figurska2008humans}, and the ``ground truth'' given by the uniformly random distribution, is presented in Figure~\ref{fig:patterns_frequency}.



Below we define each of these pattern metrics for an arbitrary sequence of numbers $x_1, x_2, \ldots, x_n$.
Denote by $\mathbb{I}$ the indicator function that takes the value of $1$ if the condition $t$ is correct and $0$ otherwise, i.e.
\[
    \mathbb{I}(t) = \left\{\begin{array}{ll}
        1 & \text{if $t$ is True},
        \\
        0 & \text{if $t$ is False}.
    \end{array}\right.
\]

The \textit{repeat frequency} is defined as the percentage of pairs of adjacent numbers that consist of the same digit:
\[
    \texttt{repeat} = \frac{\sum_{i=1}^{n-1} \mathbb{I}(x_i = x_{i+1})}{n - 1}.
\]
For a uniformly randomly distributed sequence of $10$ digits, the \textit{repeat} value should be $0.1$. For human participants it was $0.076$, and for ChatGPT we observed $0.001$.

The \textit{increase/decrease frequency} is defined as the percentage of pairs of adjacent numbers that consist of the pair of sequentially increasing/decreasing digits, respectively:
\[
    \texttt{increase/decrease} = \frac{\sum_{i=1}^{n-1} \mathbb{I}(x_i \pm 1 = x_{i+1})}{n - 1}.
\]
For a uniformly randomly distributed sequence the \textit{increase/decrease} values should be $0.09/0.09$. For human participants it was found to be $0.154/0.169$, and for ChatGPT we observed $0.063/0.078$.



\subsubsection*{Digits Frequency}
The distribution of individual digits was computed across all $10,000$ responses, defined as
\[
    \texttt{digit\_frequency} = \frac{\sum_{i=1}^{n-1} \mathbb{I}(x_i = \texttt{digit})}{n}.
\]

\begin{figure*}
    \includegraphics[width=.95\linewidth]{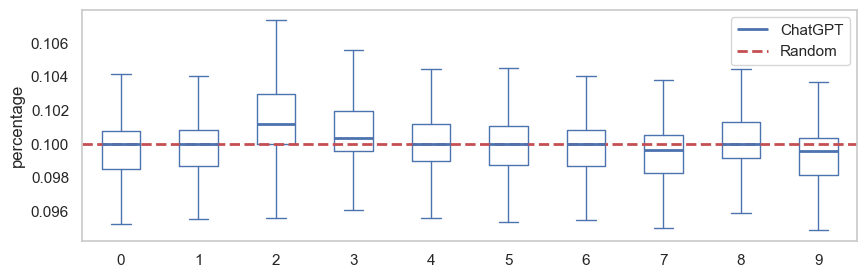}
    \caption{Distribution of individual digit frequencies across 10,000 sequences generated by ChatGPT, compared to the uniformly random distribution.}
    \Description{Distribution of individual digit frequencies in sequences generated by ChatGPT.}
    \label{fig:digits_frequency}
\end{figure*}

Ideally, each digit should appear $10\%$ of the time in uniformly random sequences. However, we did observe a slight preference for certain digits in ChatGPT's generations, with the most most frequent digit, $2$, appearing $10.3\%$ on average, and the least frequent digit, $9$, appearing $9.9\%$ on average. While these results do not exhibit perfect uniformity, they do suggest a relatively balanced distribution with only minor deviations from the ideal average. The detailed distribution is presented in Figure~\ref{fig:digits_frequency}.


\section{Discussion}

In this study, we investigate the ability of a generative pre-trained LLM to generate random sequences of digits and compare these outputs both to human-generated sequences and to sequences that are ideally random.
In particular, we test the most recent version of ChatGPT-3.5 on a simulated RNGT modeled on prior human studies~\cite{persaud2005humans, figurska2008humans} to identify any underlying similarities between sequences generated by an LLM trained predominantly on human-written data and those produced by actual humans.
Our findings indicate that while ChatGPT-generated sequences are more uniform than human-produced ones, they lack the perfect evenness characteristic of pseudorandomly generated sequences. This positions LLMs in a unique intermediate category, blending elements of both human-like variability and computational randomness.


Notably, ChatGPT significantly outperformed human capabilities in avoiding common human biases related to pattern repetition. It demonstrated a closer affinity to pseudorandom behavior in terms of the frequencies of increasing and decreasing sequences, but simultaneously showed an inclination towards fewer repeated digits than typically observed in true random generation—a tendency that, in our study, was taken to the extreme as it produced almost no consecutive duplicates at all. This aspect might suggest that within the confines of RNGTs, ChatGPT's behavior may be more influenced by its algorithmic nature rather than an emulation of the underlying human cognitive processes governing RNG.

The inclination of ChatGPT to produce fewer patterns than is commonly seen in human generations raises questions about the behavioral dynamics of GenAI systems. This characteristic could reflect a fundamental difference in how LLMs process and generate 'random' sequences compared to humans, pointing towards an inherent algorithmic bias that minimizes repetition and structure — a feature that might be rooted in the training methodologies or objective functions used in model development. Such observations align with emerging discussions in AI behavioral sciences, which probe the extent to which AI behaviors mimic human actions and where they diverge due to their computational origins \cite{lake2017building, li2023metaagents}.

Understanding these nuances is vital for advancing how we model human-AI interaction and optimize GenAI utilities to better align with human behaviors and objectives. The distinct rejection of human-pattern behaviors exhibited by ChatGPT shown in our results offers valuable insights for designing new experimental methodologies that are tailored specifically for AI behavioral research. This becomes particularly relevant when considering applications of GenAI in fields that require nuanced understanding of human behavior, such as psychology, education, and healthcare.

\subsubsection*{Technical Limitations}

In exploring the randomness and ``creativity'' of sequences generated by LLMs, it's important to consider the role of the model's temperature setting, which directly influences variability and novelty in outputs\footnote{\url{https://platform.openai.com/docs/api-reference/chat/create\#chat-create-temperature}}.
Theoretically, higher temperatures could enhance the randomness of outputs in a way similar to activating the ``creative center'' in the human brain that drives novelty~\cite{bains2008random}. However, our study intentionally refrains from altering default settings to better compare the model's natural performance under standard operating conditions rather than its maximum capability for generating random sequences.


Despite the implemented potential for the model to generate up to $922$ integers per sequence, the actual outputs generally capped at approximately $341$ digits. This discrepancy highlights a limitation in the model's ability to sustain long sequences under default settings and raises questions about the upper limits of generation in practical applications.



Furthermore, the direct translation of human RNGT prompts to LLM testing may not fully capture the nuanced differences in how humans and machines process and execute the given task, and the intentional engineering of tailored prompts that better align with the operational paradigms of LLMs would likely provide significant enhancements to performance.





\subsubsection*{Related \& Future Work}
Recent studies have employed LLMs in various capacities to predict human behavior \cite{binz2023turning}, assess cognitive abilities \cite{binz2023using}, and demonstrate traits such as reasoning and decision-making \cite{hagendorff2023human, webb2023emergent}, creativity \cite{stevenson2022putting}, and problem solving \cite{orru2023human}, with most of this research notably being performed on GPT-series models.


Our research contributes to this growing field by examining common metrics used to assess RNGT performances in human studies. As the understanding of human randomness deepens, future investigations could benefit from applying more sophisticated pattern recognition metrics to the analysis of LLM generations. Such metrics might include coupon scores \cite{ginsburg1994random}, the Evans Random Number Generation Index \cite{evans1978monitoring}, Guttmann's Null-Score Quotient \cite{brugger1996random}, and the Turning Point Index \cite{azouvi1996working}, among others \cite{ginsburg1994random, towse1998analyzing} to provide more nuanced insights into the subtleties of AI-generated randomness and its resemblance to human cognitive processes.



Expanding the range of models tested to include both proprietary (Anthropic's Claude, Google's Gemini, Cohere's Command) and open-source (Meta's LLaMA, Google's Gemma, Mistral AI's Mistral) alternatives may also allow for a greater understanding of which LLM capabilities, limitations, and trends are currently model-agnostic vs model-specific \cite{team2023gemini, touvron2023llama, team2024gemma, jiang2023mistral}.
Additionally, adjustments to the prompting strategy — such as persona-based prompts or instructions that explicitly mimic human task conditions — could further tailor how each individual model "understands" and responds to the task at hand, and adjusting parameters like \texttt{temperature}, \texttt{top\_p}, and \texttt{frequency\_penalty} could be utilized to encourage forced replication of either human- or computer-generated RNG for different research objectives.


Furthermore, there is potential in exploring the generation of random sequences using non-numeric text, such as letters or symbols. Since LLMs are predominantly trained on linguistic rather than numerical data, tasks that involve language-based randomness might align more closely with their inherent capabilities.
Such studies have previously been conducted with humans \cite{van1998age} and may offer alternative insights into whether or not LLMs may simulate human-like randomness in more abstracted contexts.
\bibliographystyle{ACM-Reference-Format}
\bibliography{references}

\end{document}